\documentclass[runningheads,a4paper]{llncs}

\usepackage{amssymb}
\usepackage{graphicx}
\usepackage{cite}
\usepackage{url}
\usepackage{fancyhdr}
\fancypagestyle{firststyle}
    {
        \fancyhf{} 
        \fancyhead[C]{\small Genetic Programming Theory \& Practice XVII, Michigan State U., May 16-19, 2019}
        \fancyfoot{}
    }
\urldef{\mailsa}\path|{sipper,jhmoore,ryanurb}@upenn.edu|    
\usepackage[leftcaption]{sidecap}
\usepackage[colorlinks = true,linkcolor = blue, urlcolor  = blue, citecolor = blue, anchorcolor = blue]{hyperref}
\newcommand{\keywords}[1]{\par\addvspace\baselineskip
\noindent\keywordname\enspace\ignorespaces#1}

\begin{document}
\mainmatter  

\title{New Pathways in Coevolutionary Computation}

\titlerunning{New Pathways in Coevolutionary Computation}
\author{Moshe Sipper\inst{1,2}%
\thanks{This work was supported by National Institutes of Health (USA) grants AI116794, LM010098, and LM012601. 
\textcopyright \,
Springer Nature Switzerland AG 2020,
W. Banzhaf et al. (eds.), Genetic Programming Theory and Practice XVII, Genetic and Evolutionary Computation, \protect\url{https://doi.org/10.1007/978-3-030-39958-0_15}. }%
\and Jason H. Moore\inst{1} \and Ryan J. Urbanowicz\inst{1}}
\authorrunning{Sipper et al.}
\institute{Institute for Biomedical Informatics, University of Pennsylvania, Philadelphia, PA 19104-6021, USA \and
Department of Computer Science, Ben-Gurion University, Beer Sheva 84105, Israel \\
\mailsa\\
\url{http://epistasis.org/}}

\toctitle{New Pathways in Coevolutionary Computation}
\tocauthor{M. Sipper, J. H. Moore, R. J. Urbanowicz}
\maketitle
\thispagestyle{firststyle}

\begin{abstract}
The simultaneous evolution of two or more species with coupled fitness---\textit{coevolution}---has been put to good use in the field of evolutionary computation. Herein, we present two new forms of coevolutionary algorithms, which we have recently designed and applied with success. 
\textit{OMNIREP} is a \textit{cooperative} coevolutionary algorithm that discovers \textit{both} a representation and an encoding for solving a particular problem of interest.
\textit{SAFE} is a \textit{commensalistic} coevolutionary algorithm that maintains two coevolving populations: a population of candidate solutions and a population of candidate objective functions needed to measure solution quality during evolution.


\end{abstract}

\section{Coevolutionary Computation}
\label{sec:coevol}
In biology, coevolution occurs when two or more species reciprocally affect each other's evolution. Darwin mentioned evolutionary interactions between flowering plants and insects in \textit{Origin of Species}. The term coevolution was coined by Paul R. Ehrlich and Peter H. Raven in 1964.\footnote{\url{https://en.wikipedia.org/wiki/Coevolution}}

\begin{figure}
\centering
\begin{tabular}{ccc}
\includegraphics[height=0.22\textwidth]{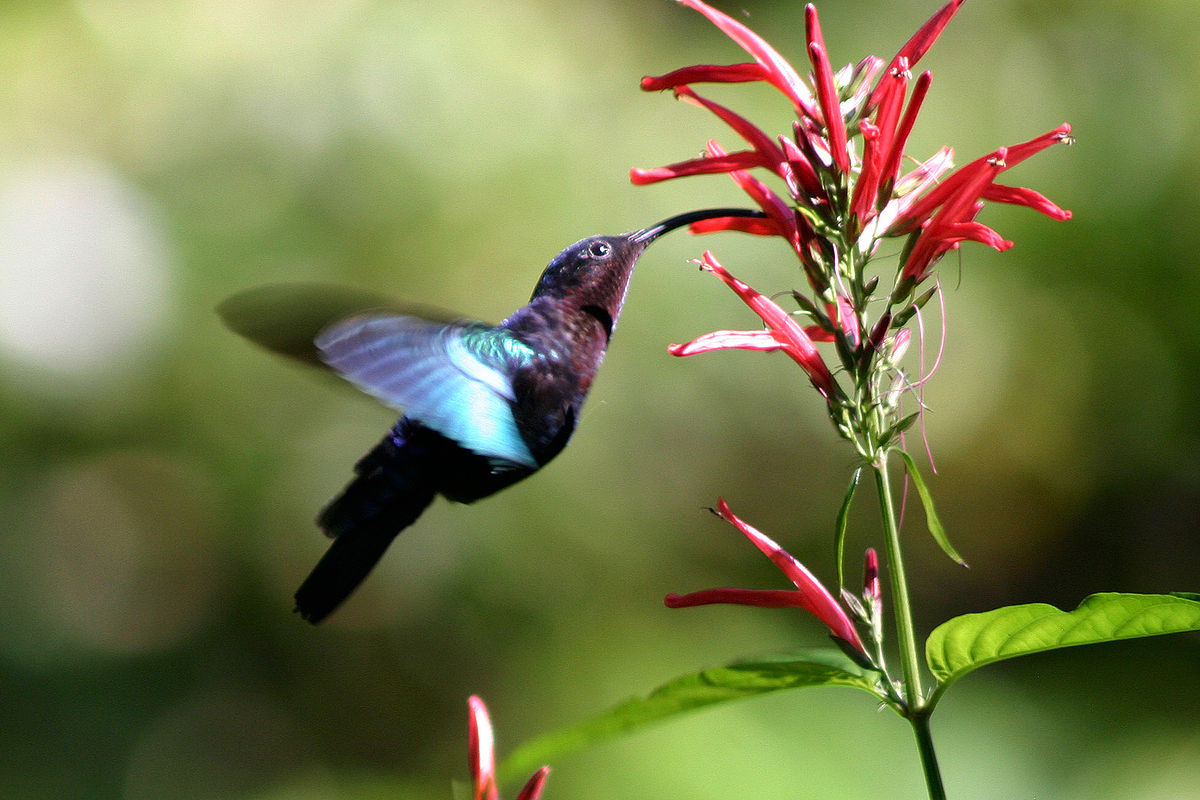} &
\includegraphics[height=0.22\textwidth]{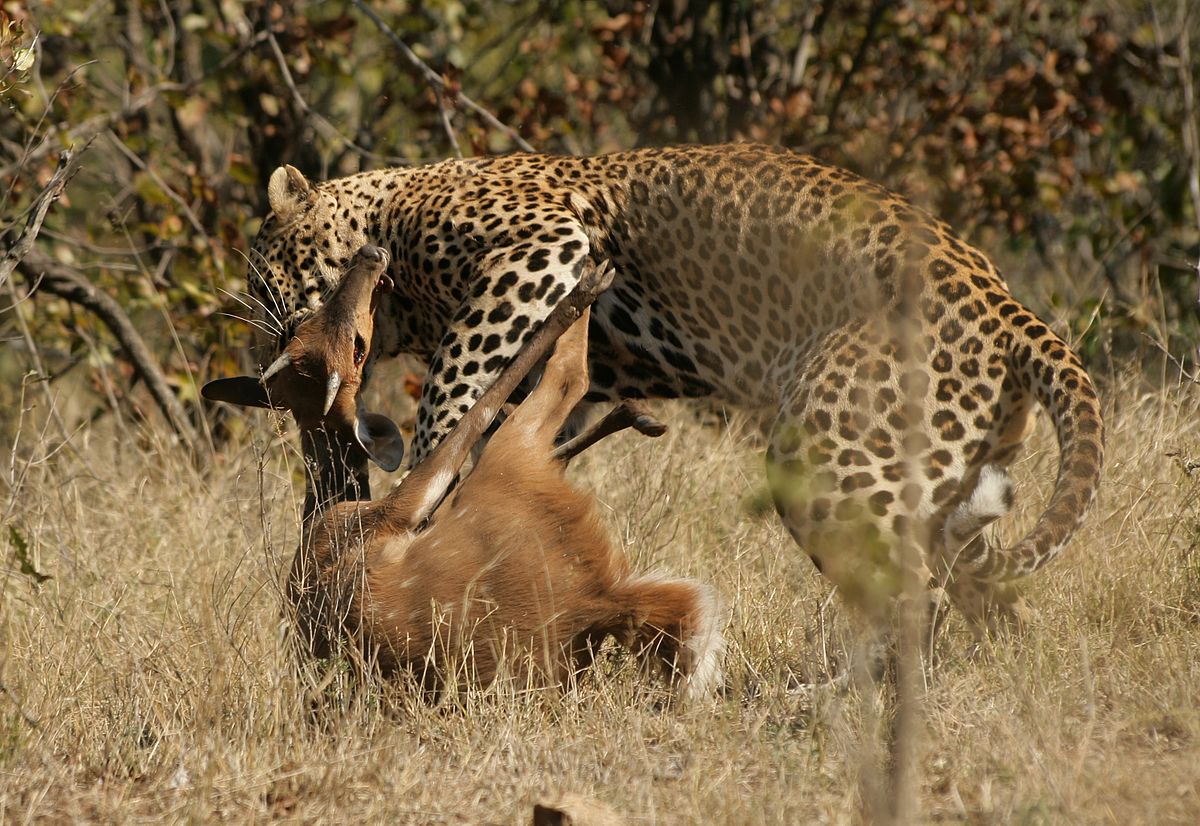} &
\includegraphics[height=0.22\textwidth]{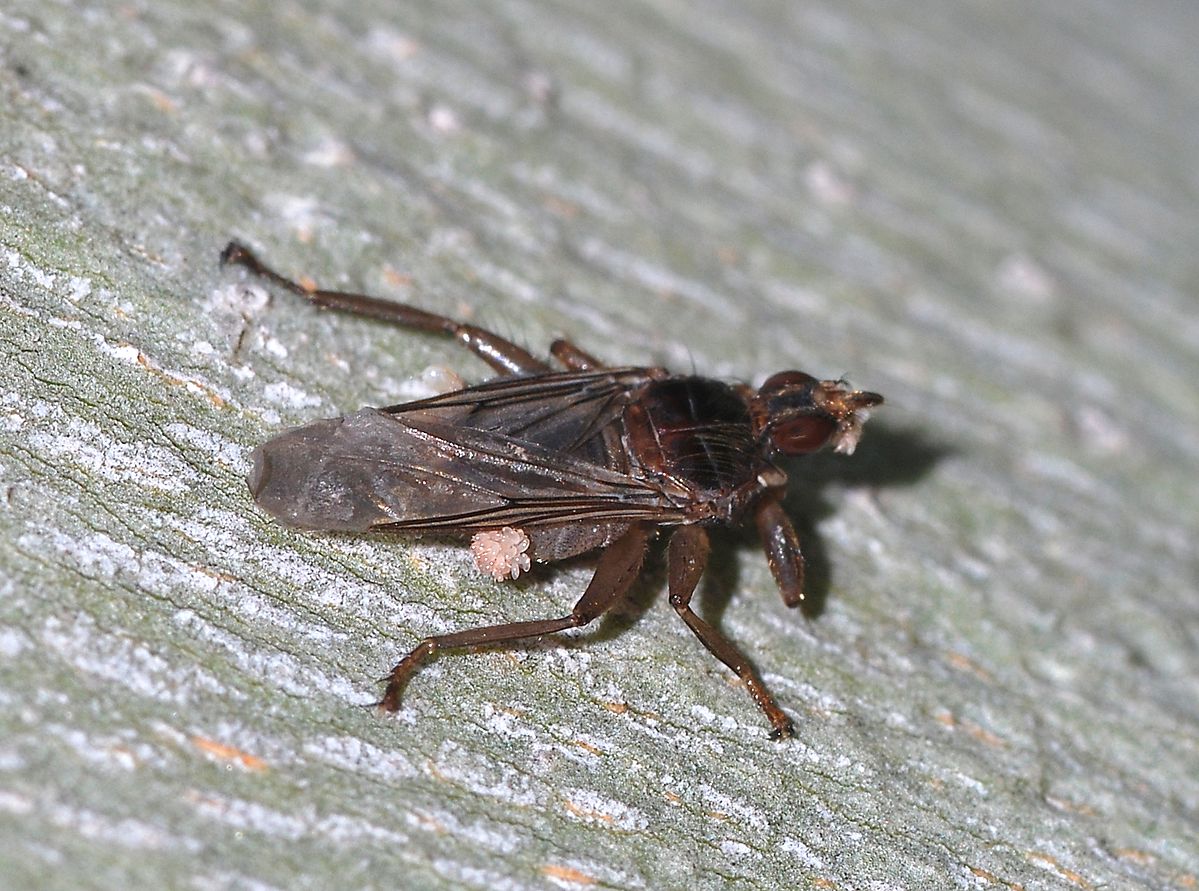} \\
(a) & (b) & (c) \\
\end{tabular}
\caption{Coevolution: 
(a) cooperative: Purple-throated carib feeding from and pollinating a flower (credit: Charles J Sharp,
\protect\url{https://commons.wikimedia.org/wiki/File:Purple-throated_carib_hummingbird_feeding.jpg});
(b) competitive: predator and prey---a leopard killing a bushbuck
(credit: NJR ZA, \protect\url{https://commons.wikimedia.org/wiki/File:Leopard_kill_-_KNP_-_001.jpg});
(c) commensalistic: Phoretic mites attach themselves to a fly for transport
(credit: Alvesgaspar, \protect\url{https://en.wikipedia.org/wiki/File:Fly_June_2008-2.jpg}).
} \label{fig:coevolution}
\end{figure}

Coevolutionary algorithms simultaneously evolve two or more populations with coupled fitness \cite{Pena:2001}. Strongly related to the concept of symbiosis, coevolution can be mutualistic (cooperative), parasitic (competitive), or commensalistic (Figure~\ref{fig:coevolution}):\footnote{\url{https://en.wikipedia.org/wiki/Symbiosis}} 
1) In cooperative coevolution, different species exist in a relationship in which each individual (fitness) benefits from the activity of the other; 2) in competitive coevolution, an organism of one species competes with an organism of a different species; and 3) with commensalism, members of one species gain benefits while those of the other species neither benefit nor are harmed.

A cooperative coevolutionary algorithm involves a number of independently evolving species, which come together to obtain problem solutions. The fitness of an individual depends on its ability to collaborate with individuals from other species \cite{zaritsky2004,Pena:2001,Potter:2000,Dick:2014}. 

In a competitive coevolutionary algorithm the fitness of an individual is based on direct competition with individuals of other species, which in turn evolve separately in their own populations. Increased fitness of one of the species implies a reduction in the fitness of the other species \cite{Hillis:1990}. 

We have recently developed two new coevolutionary algorithms, which will be reviewed herein: OMNIREP and SAFE \cite{Sipper2019robot,Sipper2019multi,Sipper2019omnirep}.

OMNIREP aims to aid in one of the major tasks faced by an evolutionary computation (EC) practitioner, namely, deciding how to represent individuals in the evolving population. This task is actually composed of two subtasks: defining a data structure that is the representation and defining the encoding that enables to interpret the representation. OMNIREP discovers \textit{both} a representation and an encoding that solve a particular problem of interest, by employing two coevolving populations. 

SAFE---Solution And Fitness Evolution---stemmed from our recently highlighting a fundamental problem recognized to confound algorithmic optimization: 
\textit{conflating} the objective with the objective function \cite{Sipper2018}. Even when the former is well defined, the latter may not be obvious.
SAFE is a commensalistic coevolutionary algorithm that maintains two coevolving populations: a population of candidate solutions and a population of candidate objective functions.
To the best of our knowledge, SAFE is the first coevolutionary algorithm to employ a form of commensalism.

We first turn to OMNIREP (Section~\ref{sec:omnirep}), followed by SAFE (Section~\ref{sec:safe}), and ending with concluding remarks (Section~\ref{sec:conc}). 
This chapter summarizes our research. For full details please refer to \cite{Sipper2019robot,Sipper2019multi,Sipper2019omnirep}.
NB: The code for both OMNIREP and SAFE is available at \url{https://github.com/EpistasisLab/}.

\section{OMNIREP}
\label{sec:omnirep}

\begin{quote}
\textit{One Representation to rule them all,  One Encoding to find them,  \\
One Algorithm to bring them all and in the Fitness bind them.\\
In the Landscape of Search where the Solutions lie.}
\end{quote}

One of the basic tasks of the EC practitioner is to decide how to represent individuals in the (evolving) population, i.e., precisely specify the genetic makeup of the artificial entity under consideration. As stated by \cite{Eiben:2003}: ``Technically, a given representation might be preferable over others if it matches the given problem better, that is, it makes the encoding of candidate solutions easier or more natural.'' 

One of the EC practitioner's foremost tasks is thus to identify a representation---a data structure---and its encoding, or \textit{interpretation}. These can be viewed, in fact, as two distinct tasks, though they are usually dealt with simultaneously. To wit, one might define the representation as a bitstring and in the same breath go on to state the encoding, e.g., ``the 120-bit bitstring represents 4 numerical values, each encoded by 30 bits, which are treated as signed floating-point values''.

\begin{figure}
    \centering
    \includegraphics[width=0.9\textwidth]{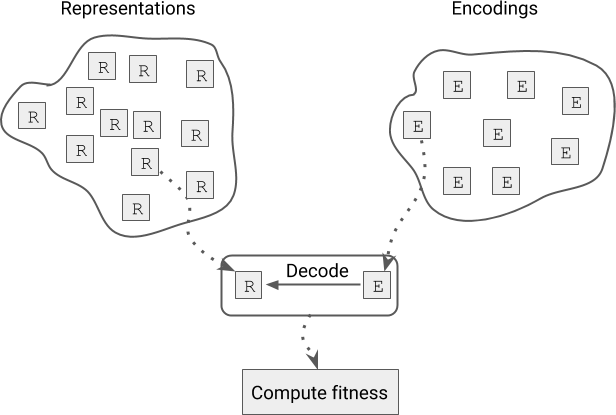}
    \caption{Fitness computation in OMNIREP, where two populations coevolve, one comprising representations, the other encodings. Fitness is computed by combining a representation individual (R) with an encoding individual (E).}
    \label{fig:OMNIREP-e}
\end{figure}

OMNIREP uses cooperative coevolution with two coevolving populations, one of representations, the other of encodings. The evolution of each population is identical to a single-population evolutionary algorithm---except where fitness is concerned (Figure~\ref{fig:OMNIREP-e}).

Selection, crossover, and mutation are performed as in a standard single-population algorithm. To compute fitness the two coevolving populations cooperate. Specifically, to compute the fitness of a single individual in one population, OMNIREP uses \textit{representatives} from the other population \cite{Pena:2001}. The representatives are selected via a greedy strategy as the 4 fittest individuals from the previous generation. When evaluating the fitness of a particular representation individual, OMNIREP combines it 4 times with the top 4 encoding individuals, computes 4 fitness values, and uses the average fitness over these 4 evaluations as the final fitness value of the representation individual. In a similar manner OMNIREP uses the average of 4 representatives from the representations population when computing the fitness of an encoding individual. 

In \cite{Sipper2019omnirep} we applied OMNIREP successfully to four problems: 
\begin{itemize}
\item \textit{Bitstring and bit count}.
Solve cubic polynomial regression problems, 
$y=ax^3+bx^2+cx+d$, where the objective was to 
find the coefficients $a,b,c,d$ for a given dataset of $x,y$ values (independent and dependent variables).
An individual in the representations population was a bitstring of length 120. An individual in the encodings population was a list of 4 integer values, each of which specified the number of bits allocated to the respective parameter ($a, b, c$, $d$) in the representation individual.

\item \textit{Floating point and precision}.
Solve regression problems, 
$y=\sum_{j=0}^{49}a_jx^{e_j}$,
where $a_j \in \mathbb{R} \cap [0,1]$, $x \in \mathbb{R} \cap [0,1]$, $e_j \in \{0,\dots,4\}$, $j=0,\dots,49$.
An individual in the representations population was a list of 50 real values $\in [0,1]$ (the coefficients $a_j$).
An individual in the encodings population was a list of 50 integer values, each specifying the precision of the respective coefficient, namely, the number of digits $d \in \{1,\ldots,8\}$ after the decimal point.

\item \textit{Program and instructions}.
Find a program that is able to emulate the output of an unknown target program.
We considered the evolution of a program composed of 10 lines, each line executing a mathematical, real-valued, univariate function, or instruction. 
The representation individual was a program
comprising 10 lines, each one executing a \textit{generic} instruction of the form \texttt{x=fi(x)}, where \texttt{fi} $\in$ \texttt{\{f1,\ldots,f5\}}. The program had one variable, \texttt{x}, which was set to a specific value \texttt{v} at the outset, i.e., to each (10-line) program, the instruction \texttt{x=v} was added as the first line. \texttt{v} was thus the program's input. After a program finished execution, its output was taken as the value of \texttt{x}. To run a program one needs to couple it with an encoding individual, which provides the specifics of what each \texttt{fi} performs. Figure~\ref{fig:sample_prog} shows an example.

\begin{SCfigure}[5]
    \centering
    \begin{tabular}{r|l}
    Representation & Encoding \\ \hline
        \texttt{x=v} \hspace{18px} & \texttt{f1:} \texttt{mul10} \\
        \texttt{x=f1(x)}           & \texttt{f2:} \texttt{fabs} \\
        \texttt{x=f2(x)}           & \texttt{f3:} \texttt{tan} \\
        \texttt{x=f3(x)}           & \texttt{f4:} \texttt{mul10} \\
        \texttt{x=f4(x)}           & \texttt{f5:} \texttt{minus2} \\
        \texttt{x=f2(x)}           &  \\
        \texttt{x=f2(x)}           &  \\
        \texttt{x=f5(x)}           &  \\
        \texttt{x=f2(x)}           &  \\
        \texttt{x=f1(x)}           &  \\
        \texttt{x=f5(x)}           &  
    \end{tabular}
    \caption{OMNIREP `program and instructions' experiment: Sample representation and encoding individuals, the former being a 10-line program with generic instructions, and the latter being the instruction meanings.}
    \label{fig:sample_prog}
\end{SCfigure}

\item \textit{Image and blocks}.
Herein, we delved into evolutionary art, 
wherein artwork is generated through an evolutionary algorithm. Our goal was to evolve images that closely matched a given target image, a ``standard of beauty'' as it were.
The representation individual's genome was a list of pixel indexes, with each index considered the start of a same-color block of pixels. 
The encoding individual was a list equal in length to the representation individual, consisting of tuples $(b_i,c_i)$, where $b_i$ was block $i$'s length, and $c_i$ was block $i$'s color. If a pixel was uncolored by any block it was assigned a default base color.
Sample evolved artwork is shown in Figure~\ref{fig:evolved-pics}.

\begin{SCfigure}[5]
    \centering
    \begin{tabular}{c@{\hskip 10px}c@{\hskip 10px}c}
     \includegraphics[height=0.13\textheight]{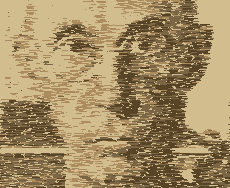} &
      \includegraphics[height=0.13\textheight]{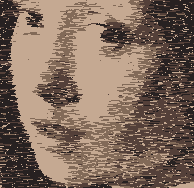} & \includegraphics[height=0.13\textheight]{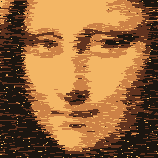}
    \end{tabular}
    \caption{Sample artwork evolved by OMNIREP.}
    \label{fig:evolved-pics}
\end{SCfigure}

\end{itemize}

OMNIREP was able to solve all problems successfully. Moreover, it usually found better encodings (e.g., more compact---using less bits or less precision) than fixed-representation schemes, with no degradation in performance. 
For full details see \cite{Sipper2019omnirep}.

\section{SAFE}
\label{sec:safe}

We have recently highlighted a fundamental problem recognized to confound algorithmic optimization, namely, \textit{conflating} the objective with the objective function \cite{Sipper2018}. Even when the former is well defined, the latter may not be obvious. We presented an approach to automate the means by which a good objective function might be discovered, through the introduction of SAFE---Solution And Fitness Evolution---a commensalistic coevolutionary algorithm that maintains two coevolving populations: a population of candidate solutions and a population of candidate objective functions \cite{Sipper2019robot,Sipper2019multi}. 

\begin{figure}
\centering
\begin{tabular}{c@{\hskip 20px}c}
\includegraphics[width=0.4\textwidth]{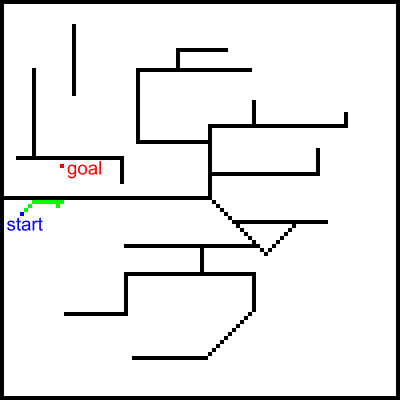} &
\includegraphics[width=0.4\textwidth]{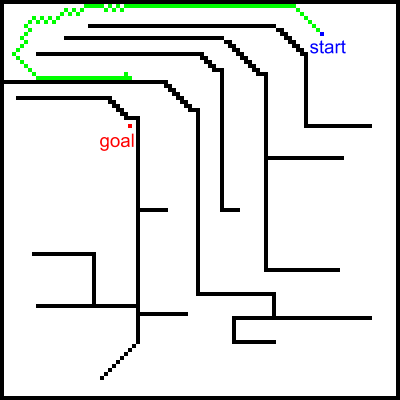} \\
\texttt{maze1} & \texttt{maze2}
\end{tabular}
\caption{In a maze problem a robot begins at the start square and must make its way to the goal square (objective). Shown above are paths (green) of robots evolved by a standard evolutionary algorithm with fitness measured as distance-to-goal, evidencing how conflating the objective with the objective function leads to a non-optimal solution.} \label{fig:maze-ea}
\end{figure}

Consider a robot navigating a maze, wherein the challenge is to evolve a robotic controller such that the robot, when placed in the start position, is able to make its way to the goal. 
It seems intuitive that the fitness of a given robotic controller be defined as a function of the distance from the robot to the objective, as done, e.g., by \cite{Lehman2008}. However, reaching the objective may be difficult since the robot is faced with a deceptive landscape, where higher fitness (i.e., being reasonably close to the goal) may not imply that the robot is ``almost there''. It is quite easy for the robot to attain a fairly good fitness value, yet be stuck behind a wall in a local optimum---quite far from the objective in terms of the path needed to be taken. Indeed, our experiments with such a fitness-based evolutionary algorithm \cite{Sipper2019robot} produced the expected failure, demonstrated in Figure~\ref{fig:maze-ea}.

One solution to this conflation problem was offered by \cite{Lehman2008} in the form of novelty search, which \textit{ignores} the objective and searches for novelty. However, novelty for the sake of novelty alone lacks incentive for solutions that reach and stay at the objective.

Perhaps, though, the issue lies with our ignorance of the \textit{correct} objective function. That is the motivation behind the SAFE algorithm.

\begin{figure}
\centering
\includegraphics[width=0.95\textwidth]{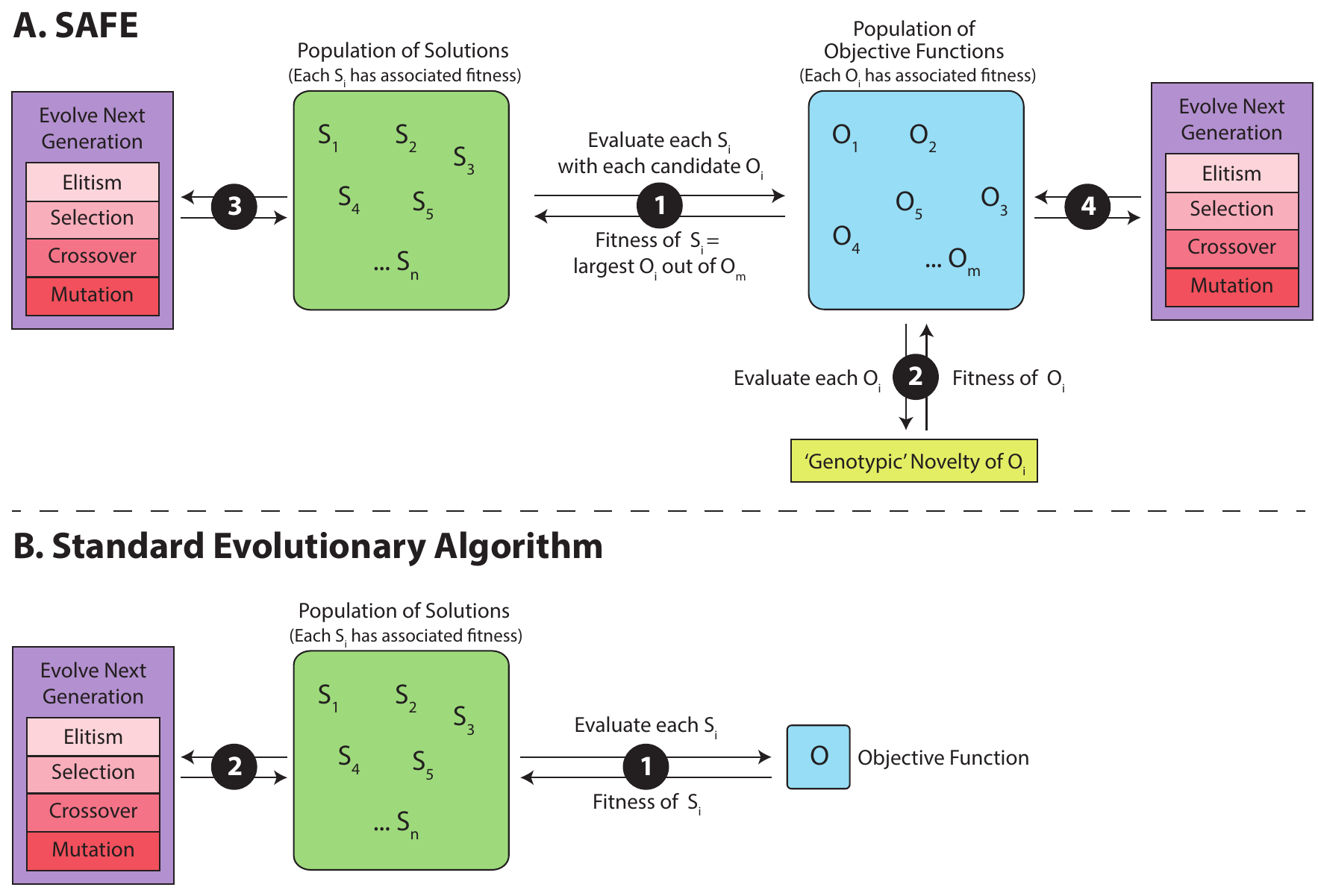} 
\caption{A single generation of SAFE vs. a single generation of a standard evolutionary algorithm. The numbered circles identify sequential steps in the respective algorithms. The objective function can comprise a single or multiple objectives.} \label{fig:safe}
\end{figure}

SAFE is a coevolutionary algorithm that maintains two coevolving populations: a population of candidate solutions and a population of candidate objective functions.
The evolution of each population is identical to a standard, single-population evolutionary algorithm---except where fitness computation is concerned, as shown in Figure~\ref{fig:safe}.

We applied SAFE to two domains: evolving robot controllers to solve mazes \cite{Sipper2019robot} and multiobjective optimization \cite{Sipper2019multi}.

Applying SAFE within the robotic domain, 
an individual in the solutions population was a list of 16 real values, representing the robot's control vector (``brain''). The controller determined the robot's behavior when wandering the maze, with its
phenotype taken to be the final position, or endpoint. The endpoint was used to compute standard distance-to-goal fitness and to compute \textit{phenotypic} novelty: compare the endpoint to all endpoints of current-generation robots \textit{and} to all endpoints in an archive of past individuals whose behaviors were highly novel when they emerged.
The final novelty score was then the average of the 15 nearest neighbors.

An individual in the objective-functions population was a list of 2 real values $[a,b]$, each $\in [0,1]$.

Every solution individual was scored by every candidate objective-function individual in the current population (Figure~\ref{fig:safe}A).
Candidate SAFE objective functions  incorporated both `distance to goal' (the evolving $a$ parameter) as well as phenotypic novelty (the evolving $b$ parameter) in order to calculate solution fitness, weighting the two objectives in a simple linear fashion. 
The best (highest) of these objective-function scores was then assigned to the individual solution as its fitness value.

As for the objective-functions population, determining the quality of an evolving objective function posed a challenge. 
Eventually we turned to a commensalistic coevolutionary strategy, where the objective functions' fitness did not depend on the population of solutions. Instead, it relied on \textit{genotypic} novelty, based on the objective-function individual's two-valued genome, $[a,b]$. The distance between two objective functions was simply the Euclidean distance of their genomes. Each generation, every candidate objective function was compared to its cohorts in the current population of objective functions and to an archive of past individuals whose behaviors were highly novel when they emerged. The novelty score was the average of the distances to the 15 nearest neighbors, and was used in computing objective-function fitness. 

\begin{figure}
\centering
\begin{tabular}{c@{\hskip 20px}c}
\includegraphics[width=0.4\textwidth]{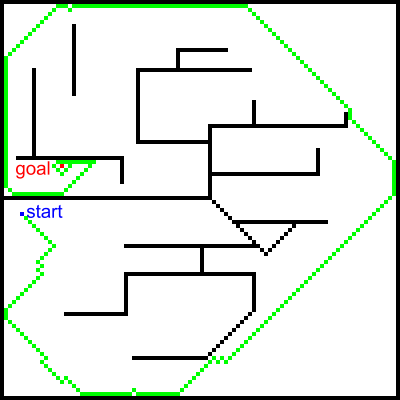} &
\includegraphics[width=0.4\textwidth]{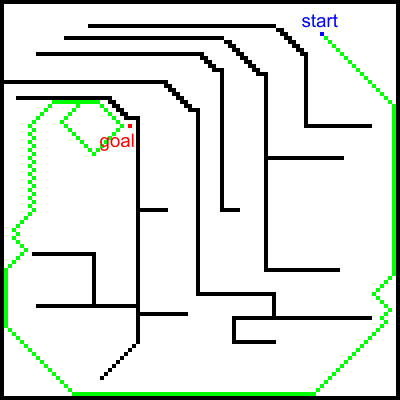} \\
\texttt{maze1} & \texttt{maze2}
\end{tabular}
\caption{Solutions to the maze problems, evolved by SAFE.} \label{fig:maze-solutions}
\end{figure}

SAFE performed far better than random search and a standard fitness-based evolutionary algorithm, and compared favorably with novelty search. 
Figure~\ref{fig:maze-solutions} shows sample solutions found by SAFE (contrast this with the  standard evolutionary algorithm, which always got stuck in a local minimum, as exemplified in Figure~\ref{fig:maze-ea}). For full details see \cite{Sipper2019robot}.

The second domain we applied SAFE to was multiobjective optmization \cite{Sipper2019multi}.
A multiobjective optimization problem involves two or more objectives all of which need to be optimized. 
Applications of multiobjective optimization abound in numerous domains \cite{zhou2011multiobjective}.

With a multiobjective optimization problem there is usually no single-best solution, but rather the goal is to identify a set of `non-dominated' solutions that represent optimal tradeoffs between multiple objectives---the \textit{Pareto front}. Usually, a representative subset will suffice.

Specifically, we applied SAFE to the solution of 
the classical ZDT problems, which
epitomize the basic setup of multiobjective optimization
\cite{zitzler2000comparison,huband2006review}.
For example, ZDT1 is defined as:
\[ f_1(\textbf{x}) = x_1 \, ,\]
\[ g(\textbf{x}) = 1 + 9/(k-1)\sum_{i=2}^{k} x_i \, ,\]
\[ f_2(\textbf{x}) = 1-\sqrt{f_1/g} \, .\]
The two objectives are to minimize both $f_1(\textbf{x})$ and $f_2(\textbf{x})$.
The dimensionality of the problem is $k=30$, i.e.,
solution vector 
$\textbf{x} = x_1,\ldots,x_{30}$, $x_i \in [0,1]$. 
The utility of this suite is that the ground-truth optimal Pareto front can be computed and used to determine and compare multiobjective algorithm performance. 

SAFE maintained two coevolving populations. An individual in the solutions population was a list of 30 real values.
An individual in the objective-functions population was a list of 2 real values $[a,b]$, each in the range $[0,1]$, defining a candidate set of weights, balancing the two objectives of the ZDT functions:
$a$ determined $f_1$'s weighting and $b$ determined $f_2$'s weighting. 

Note that, as opposed to many other multiobjective optimizers, SAFE did not rely on measures of the Pareto front (i.e., a Pareto front was not employed to calculate solution fitness, or as a standard for selecting parent solutions to generate offspring solutions).

We tested SAFE on four ZDT problems---ZDT1, ZDT2, ZDT3, ZDT4---recording the evolving Pareto front as evolution progressed. 
We compared our results with two very recent studies
by Cheng et al. \cite{cheng2017novel} and 
by Han et al. \cite{han2018improved},
finding that SAFE was able to perform convincingly better on 3 of the 4 problems.
For full details see \cite{Sipper2019multi}.

\section{Concluding Remarks}
\label{sec:conc}
The experimentation performed to date is perhaps not definitive yet but we hope to have offered at least proof-of-concept of our two new coevolutionary algorithms. Both have been shown to be successful in a number of domains.

There are several avenues of future research that present themselves, including:
\begin{itemize}
\item Study and apply both algorithms to novel domains. We have been looking into applying SAFE to datasets created by the GAMETES system, which models epistasis \cite{urbanowicz2012gametes}. We have also created additional art by devising novel encoding-representation couplings for OMNIREP (Figure~\ref{fig:more-art}).

\begin{figure}
\centering
\includegraphics[width=\textwidth]{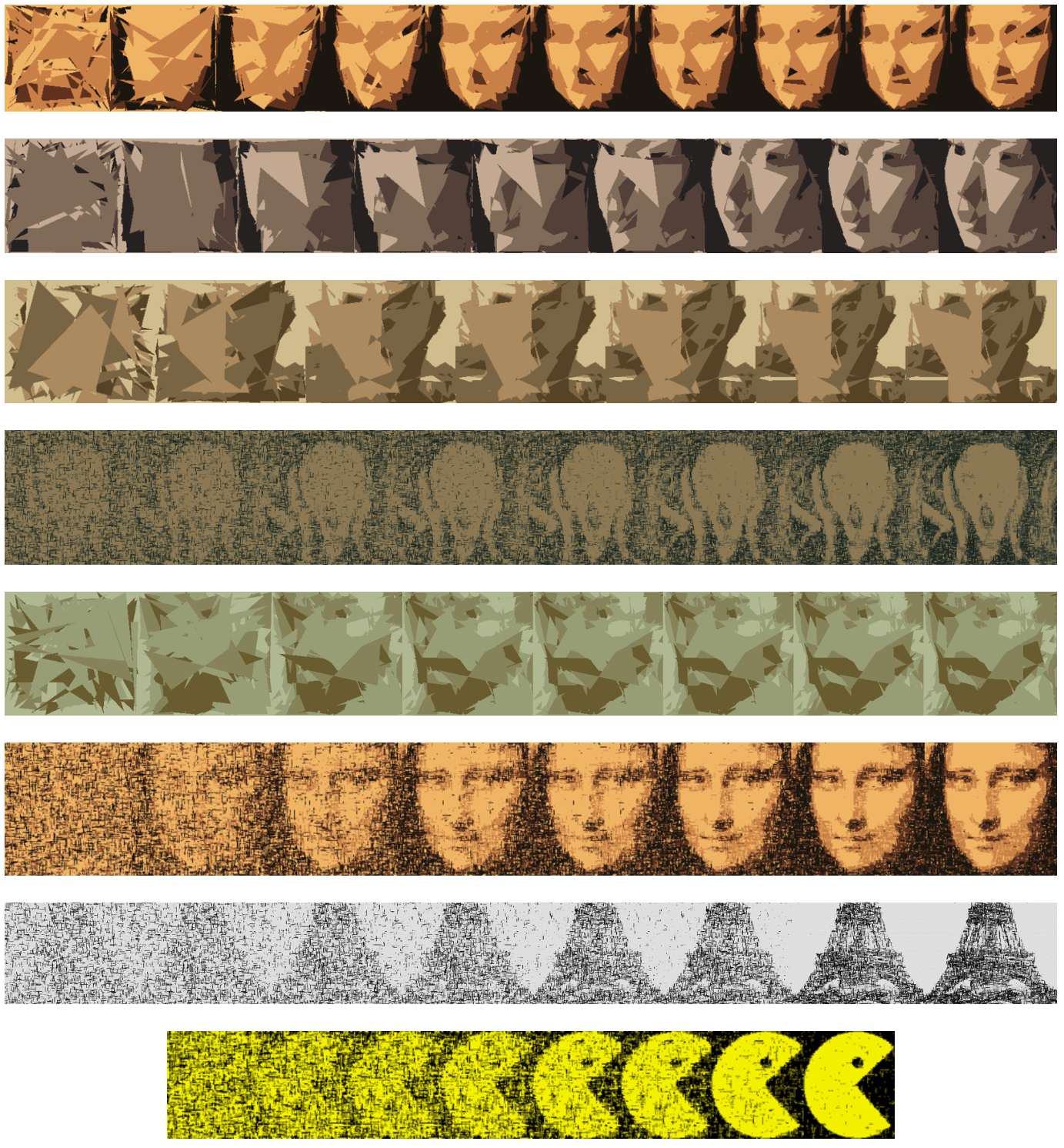}
\caption{Additional artwork created by OMNIREP using novel encoding-representation couplings (involving polygons, and horizontal and vertical blocks). Each row shows a single evolutionary run, from earlier generations (left) to later generations (right).} \label{fig:more-art}
\end{figure}

\item Study the coevolutionary dynamics engendered by OMNIREP and SAFE.

\item Cooperative or competitive versions of SAFE (which is currently commensalistic), i.e., finding ways in which the objective-function population depends on the solutions population.

\item Examine the incorporation of more sophisticated evolutionary algorithm components (e.g., selection, elitism, genetic operators).  
\end{itemize}

\bibliographystyle{spmpsci}
\bibliography{bibfile}

\end{document}